\pdfoutput=1

\documentclass[11pt]{article}

\usepackage[]{EMNLP2022}

\usepackage{times}
\usepackage{latexsym}
\usepackage{graphicx,multirow,array,booktabs,url, subfigure}

\usepackage[T1]{fontenc}

\usepackage[utf8]{inputenc}

\usepackage{microtype}

\usepackage{inconsolata}

\usepackage{amsmath}
\usepackage{enumitem}
\usepackage[dvipsnames]{xcolor}
\usepackage{moreverb}

\usepackage{algpseudocode}
\usepackage{booktabs}
\usepackage[linesnumbered,ruled,vlined]{algorithm2e}

\usepackage{todonotes} 

\newcommand{\acronym}{\texttt{DANLI}}

\newcommand{\addafterrebuttal}[1]{#1}

\title{DANLI: Deliberative Agent for Following Natural Language Instructions}

\author{Yichi Zhang \hspace{20pt} Jianing Yang \hspace{20pt} Jiayi Pan \hspace{20pt} Shane Storks \hspace{20pt} Nikhil Devraj \\
{\bf Ziqiao Ma \hspace{20pt} Keunwoo Peter Yu \hspace{20pt} Yuwei Bao \hspace{20pt} Joyce Chai} \\
Computer Science and Engineering Division, University of Michigan \\
  \texttt{zhangyic@umich.edu}
}

\begin{document}
\maketitle
\begin{abstract}
Recent years have seen an increasing amount of work on embodied AI agents that can perform tasks by following human language instructions. However, most of these agents are {\em reactive}, meaning that they simply learn and imitate behaviors encountered in the training data. These reactive agents are insufficient for long-horizon complex tasks. 
To address this limitation, we propose a neuro-symbolic {\em deliberative} agent that, while following language instructions, proactively applies reasoning and planning based on its neural and symbolic representations acquired from past experience (e.g., natural language and egocentric vision). 
We show that our deliberative agent achieves greater than 70\% improvement over reactive baselines on the challenging TEACh benchmark. 
Moreover, the underlying reasoning and planning processes, together with our modular framework, offer impressive transparency and explainability to the behaviors of the agent. This enables an in-depth understanding of the agent's capabilities, which shed light on challenges and opportunities for future embodied agents for instruction following. \addafterrebuttal{The code is available at \url{https://github.com/sled-group/DANLI}.}


  
\end{abstract}

\vspace{-5pt}
\section{Introduction}

\vspace{-5pt}

Natural language instruction following with embodied AI agents~\cite{chai2018language, mattersim,thomason:arxiv19,reverie,shridhar2020alfred,padmakumar2021teach} is a notoriously difficult problem, where an agent must interpret human language commands to perform actions in the physical world and achieve a goal.
Especially challenging is the hierarchical nature of everyday tasks,\footnote{For example, \textit{making breakfast} may require preparing one or more dishes (e.g., toast and coffee), each of which requires several sub-tasks of navigating through the environment and manipulating objects (e.g., finding a knife, slicing bread, cooking it in the toaster), and even more fine-grained primitive actions entailed by them (e.g., walk forward, pick up knife).}
which often require reasoning about subgoals and reconciling them with the world state and overall goal. 
%
However, despite recent progress, past approaches are typically \textit{reactive}~\cite{Wooldridge1995Conceptual} in their execution of actions: conditioned on the rich, multimodal inputs from the environment, they perform actions directly without using an explicit representation of the world to facilitate grounded reasoning and planning \cite{pashevich2021episodic,zhang2021hitut,sharma2021skill}.
Such an approach is inefficient, as natural language instructions often omit trivial steps that a human may be assumed to already know \cite{zhou2021hierarchical}. Besides, the lack of any explicit symbolic component makes such approaches hard to interpret, especially when the agent makes errors.



Inspired by previous work toward \textit{deliberative} agents in robotic task planning, which apply long-term action planning over known world and goal states \cite{she14sigdial, pmlr-v164-agia22a,srivastava2021behavior,wang2022generalizable}, we introduce \acronym{}, a neuro-symbolic {\em D}eliberative {\em A}gent for following {\em N}atural {\em L}anguage {\em I}nstructions.
\acronym{} combines learned symbolic representations of task subgoals and the surrounding environment with a robust symbolic planning algorithm to execute tasks.
First, we build a uniquely rich semantic spatial representation~(Section~\ref{sec:map}), acquired online from the surrounding environment and language descriptions to capture symbolic information about object instances and their physical states.
To capture the highest level of hierarchy in tasks, we propose a neural task monitor~(Section~\ref{sec:subgoal}) that learns to extract symbolic information about task progress and upcoming subgoals from the dialog and action history.
Using these elements as a planning domain, we lastly apply an online planning algorithm~(Section~\ref{sec:plan}) to plan low-level actions for subgoals in the environment, taking advantage of \acronym{}'s transparent reasoning and planning pipeline to detect and recover from errors.

Our empirical results demonstrate that our deliberative \acronym{} agent outperforms reactive approaches with better success rates and overwhelmingly more efficient policies on the challenging Task-driven Embodied Agents that Chat (TEACh) benchmark~\cite{padmakumar2021teach}. 
Importantly, due to its interpretable symbolic representation and explicit reasoning mechanisms, our approach offers detailed insights into the agent's planning, manipulation, and navigation capabilities. 
This gives the agent a unique self awareness about the kind of exceptions that have occurred, and therefore makes it possible to adapt strategies to cope with exceptions and continually strengthen the system.

\vspace{-2pt}
\section{Problem Definition}

\vspace{-5pt}
The challenge of hierarchical tasks is prominent in the recent Task-driven Embodied Agents that Chat (TEACh) benchmark for this problem \cite{padmakumar2021teach}. 
Here, language instructions are instantiated as a task-oriented dialog between the agent and a commander (who has comprehensive knowledge about the task and environment, but cannot perform any actions), with varying granularity and completeness of guidance given.
We focus on the Execution from Dialog History (EDH) setting in TEACh, where the agent is given a dialog history as input, and is expected to execute a sequence of actions and achieve the goal set out by the commander. This setting allows us to abstract away the problem of dialog generation, and focus on the already difficult problem of instruction following from task-oriented dialog.


\begin{figure}[t!]
    \centering
    \includegraphics[width=0.87\columnwidth]{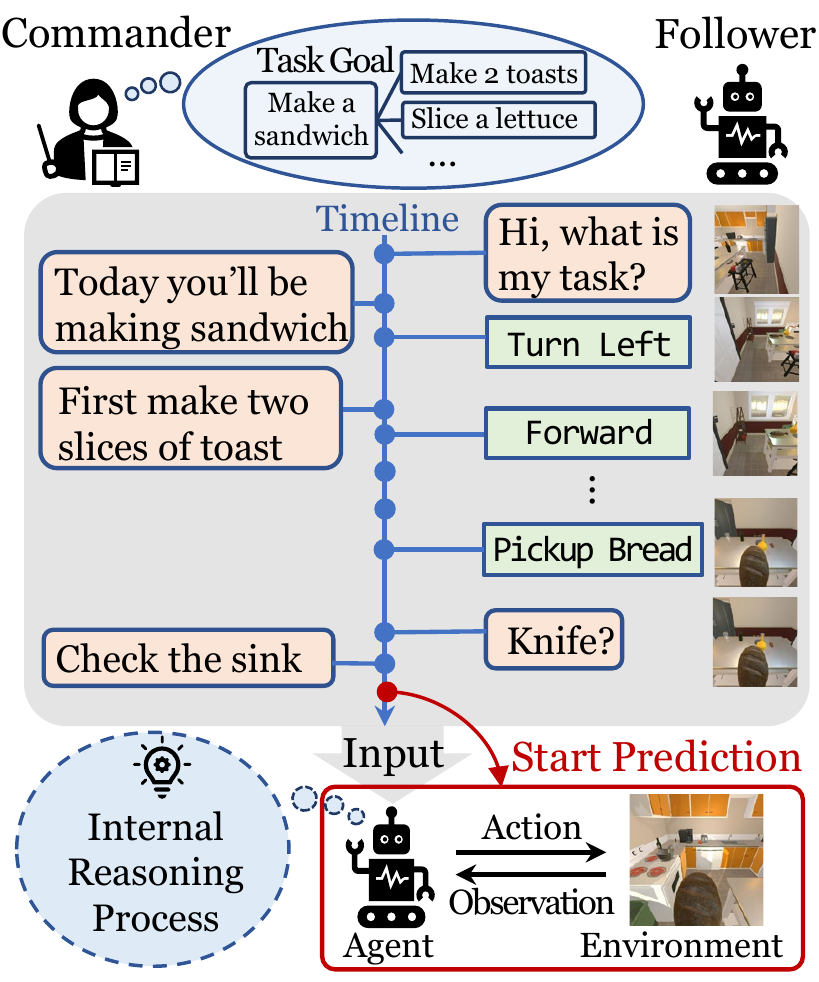}
    \vspace{-5pt}
    \caption{An example task in TEACh.} 
    \label{fig:intro}
    \vspace{-10pt}
\end{figure}

As shown in Figure~\ref{fig:intro}, a \textit{task}, e.g., \textit{Make a Sandwich}, may have several subtasks that the agent must achieve in order to satisfy the overall task goal. The success of a task/subtask is achieved by meeting a set of \textit{goal conditions}, such as slicing a bread and toasting two pieces.
At each timestep, the agent receives an egocentric visual observation of the world and the full dialog history up to that time, and may execute a single low-level \textit{action}. Actions can either involve navigation, e.g., to step forward, or manipulation, e.g., to pick up an object. Manipulation actions additionally require the agent to identify the action's target object by specifying a pixel in its field of view to highlight the object.
The execution continues until the agent predicts a \texttt{Stop} action, otherwise the session will terminate after 1000 timesteps or 30 failed actions. 
At this time, we can evaluate the agent's completion of the task.

It is worth noting that while we focus on TEACh, our approach is largely transferable between benchmark datasets and simulation environments, albeit requiring retraining of some components.

\vspace{-2pt}

\section{A Neuro-Symbolic Deliberative Agent}


\begin{figure*}[t!]
    \centering
    \includegraphics[width=0.98\textwidth]{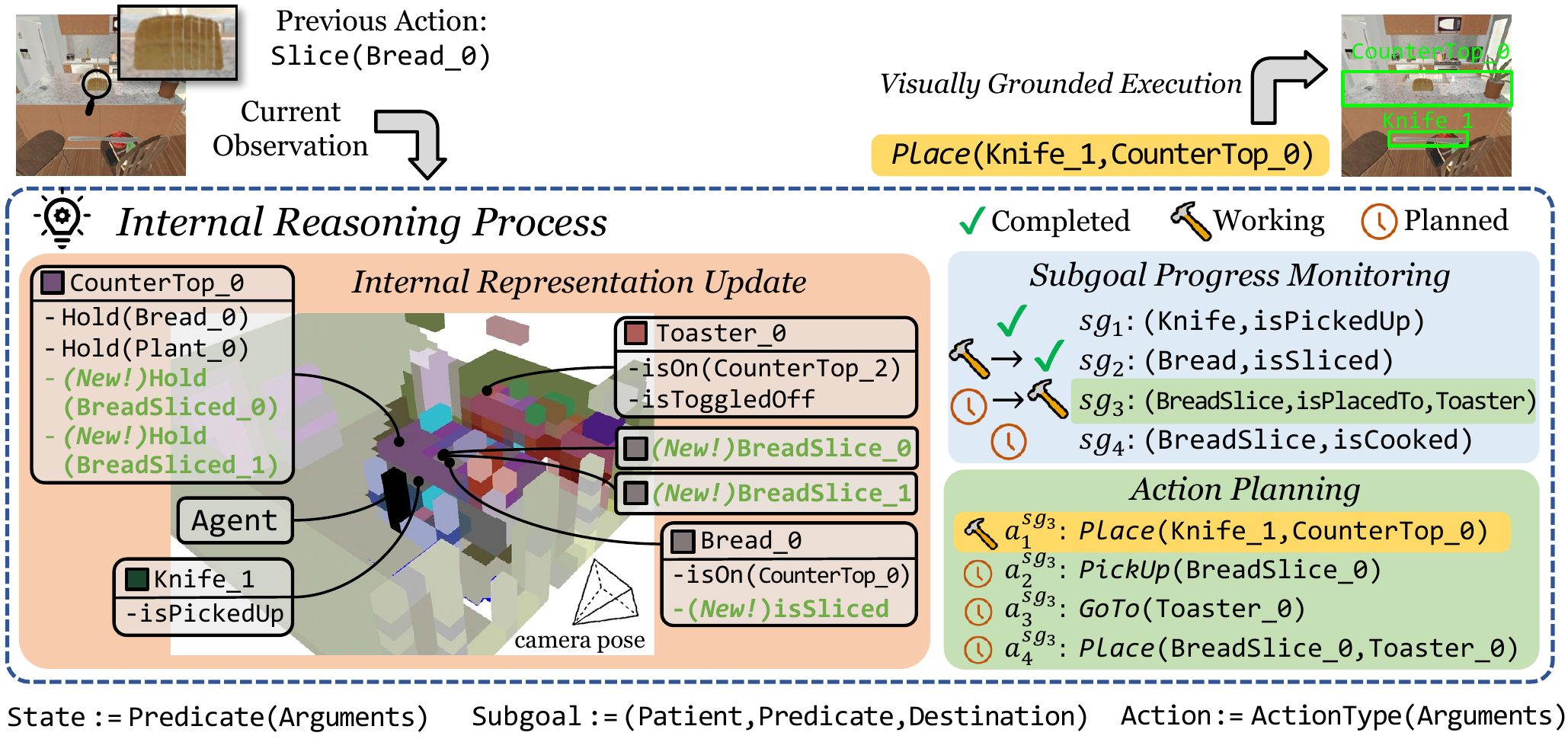}
    \vspace{-5pt}
    \caption{Illustration of our agent's reasoning process behind a single decision step. After receiving the current observation, the agent first updates its \textcolor{YellowOrange}{internal representation} (\textcolor{YellowOrange}{orange}), then checks the current \textcolor{Cyan}{subgoal progress} (\textcolor{Cyan}{blue}), and \textcolor{ForestGreen}{plans for the next steps} (\textcolor{ForestGreen}{green}). Finally the first action in the plan is popped out and grounds to the agent's ego-centric view for execution. In the pop-up boxes we show example object instances with their instance ids, states and positions in the 3D map. New instances and state changes are labeled in green. The status for each subgoal and action is labeled in front of it, where the arrows denote status transitions. }
    \label{fig:new overview}
    \vspace{-10pt}
\end{figure*}

\vspace{-5pt}
An overview of our neuro-symbolic deliberative agent is shown in Figure~\ref{fig:new overview}. We first introduce the symbolic notions used in our system. We use the object-oriented representation \cite{diuk2008object} to represent the symbolic world state. Each \textit{object instance} is assigned an instance ID consisting of its canonicalized class name and ordinal. We define a \textit{state} in the form of \texttt{Predicate(Arguments)} as an object's physical state or a relation to another object. We define \textit{subgoals} as particular states that the agent should achieve while completing the task, represented by a symbolic form \texttt{(Patient, Predicate,  Destination)}\footnote{\texttt{isPlacedTo} is the only predicate with a \texttt{Destination}.}, 
where the \texttt{Patient} and \texttt{Destination} are object classes, and  
the \texttt{Predicate} is a state that can be applied to the \texttt{Patient}.
We define an \textit{action} in the agent's plan as \texttt{ActionType(Arguments)} where each argument is an object instance.

To complete tasks, our agent reasons over a learned spatial-symbolic map representation (Section~\ref{sec:map}) to generate a hierarchical plan. At the high level, it applies a neural language model to the dialog and action history to predict the complete sequence of completed and future subgoals in symbolic form (Section~\ref{sec:subgoal}). For each predicted subgoal, it then plans a sequence of low-level actions using both the symbolic subgoal and world representations online, with robustness to various types of planning and execution failures (Section~\ref{sec:plan}).
Next, we describe how each component works and highlight our key innovations.

\vspace{-5pt}
\subsection{World Representation Construction}

\label{sec:map}
The reasoning process of an embodied AI agent relies heavily on a strong internal representation of the world. As shown in Figure \ref{fig:new overview}, we implement the internal representation as a semantic map incorporating rich symbolic information about object instances and their physical states. We introduce our methods for the construction of this representation in this section. 

\vspace{-5pt}
\paragraph{3D Semantic Voxel Map}
As the agent moves through the environment while completing a task, it constructs a 3D semantic voxel map to model its spatial layout. Following \citet{blukis2022hlsm}, we use a depth estimator to project the pixels of egocentric observation images and detected objects to 3D point cloud and bin the points into $0.25$m$^3$ voxels. The resulting map can help symbolic planner (Section \ref{sec:plan}) break down high-level navigation actions, such as \texttt{GOTO Knife\_0}, to atomic navigation actions such as \texttt{Forward, TurnLeft, LookUp}.\footnote{See Appendix \ref{sec:appendix_map} for more details on path planning.}

\vspace{-5pt}
\paragraph{Object Instance Lookup Table}
Everyday tasks can involve multiple instances of the same object, and thus modeling only object class information may be insufficient.\footnote{\addafterrebuttal{For example, when making a sandwich, the agent will likely need to distinguish the top and bottom pieces of bread to make the sandwich complete.}} As shown in the internal representation update part of Figure~\ref{fig:new overview}, we store object instance information for a single task episode in a symbolic lookup table, where each instance in the environment is assigned a unique ID once observed. These symbols in the lookup table become the planning domain of the symbolic planner (Section \ref{sec:plan}). To collect this symbolic lookup table, we use a panoptic segmentation model\footnote{As opposed to a semantic segmentation model as used in prior work \cite{chaplot2020object,min2021film,blukis2022hlsm}, which can only detect object class information.} to detect all object instances in the current 2D egocentric visual frame. These 2D instance detections are then projected into the 3D map, and we use each instance's 3D centroid and size information to match and update existing object instances' information in the lookup table\footnote{To perform this update and decide whether a newly detected instance should be merged with an existing instance or added as a new one, we use a matching algorithm described in Appendix~\ref{sec:appendix_instance_match}.}. As the agent moves through the scene and receives more visual observations, the symbolic lookup table becomes more complete and accurate.

\vspace{-5pt}
\paragraph{Physical State Prediction}






Additionally, tasks can hinge upon the physical states of particular object instances. For example, when making coffee, the agent should disambiguate dirty and clean coffee mugs and make sure to use the clean mug. 
To recognize the physical state of each object instance, we propose a physical state classification model where inputs include the image region of a detected object instance and its class identifier, and the output is physical state labels for the instance. 
As classifying the physical state from visual observation alone can introduce errors, we also incorporate the effect of the agent's actions into physical state classifications. For example, the \texttt{isToggledOn} attribute is automatically modified after the agent applies the \texttt{ToggleOn} action, overriding the classifier's prediction. 

\vspace{-5pt}
\subsection{Subgoal-Based Task Monitoring}
\vspace{-5pt}
\label{sec:subgoal}

Due to the hierarchical nature of tasks, natural language instructions may express a mix of high-level and low-level instructions. In order to monitor and control the completion of a long-horizon task given such complex inputs, we first model the sequence of high-level \textit{subgoals}, i.e., key intermediate steps necessary to complete it. 

\begin{figure}[t!]
	\centering
    \includegraphics[width=0.99\columnwidth]{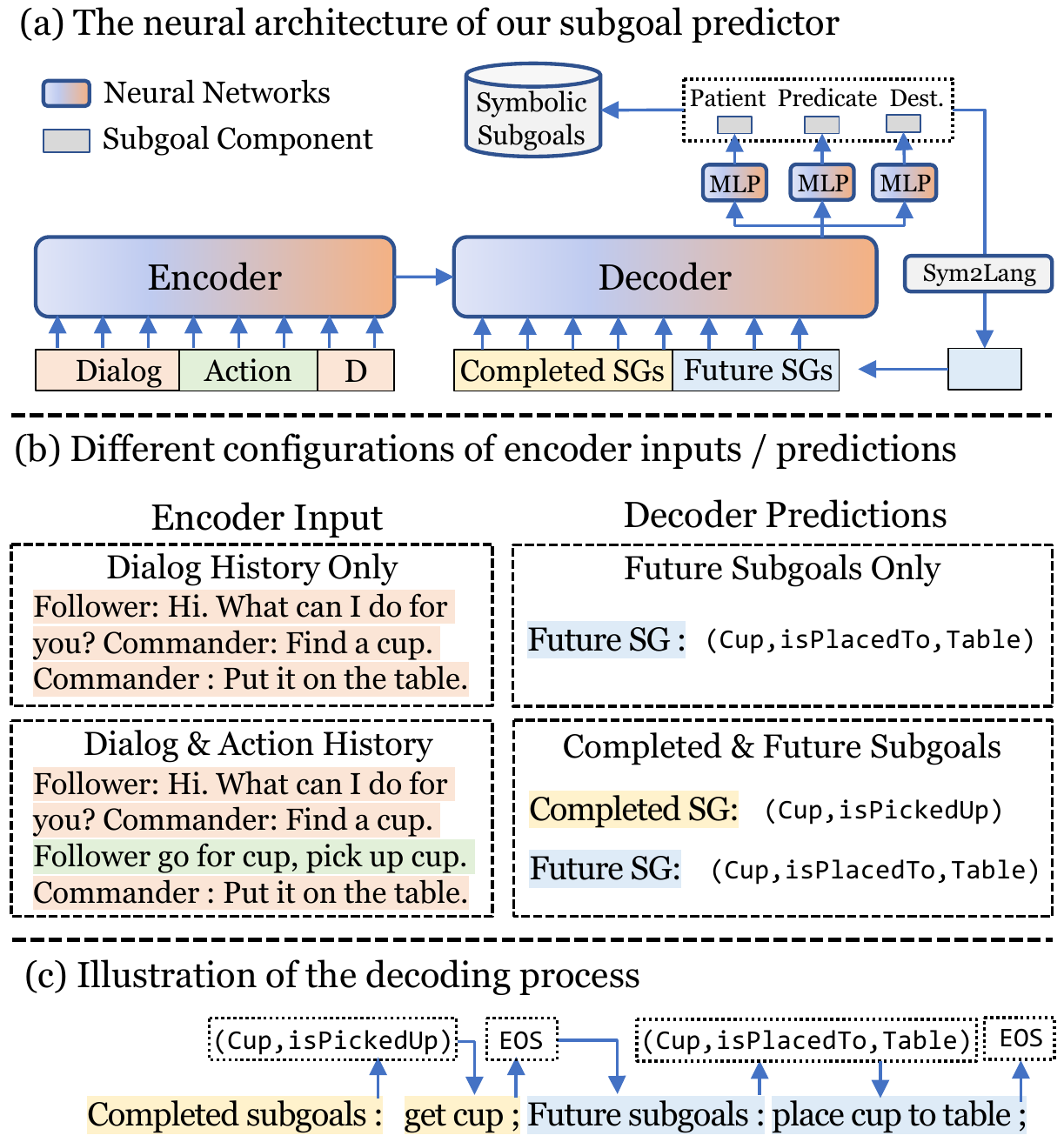}
    \caption{Overview of the subgoal learning process. Figure (a) shows the model architecture. Figure (b) shows the different input/output configurations we experiment with. Figure (c) illustrates the decoding process. We first predict completed subgoals and then predict the future subgoals conditioned on them, where we use different prompts to distinguish different types of subgoals. }
	\label{fig:subgoal_learning}
	\vspace{-10pt}
\end{figure}

As shown in Figure~\ref{fig:subgoal_learning}, we apply a sequence-to-sequence approach powered by language models to learn subgoals from the dialog and action history. At the beginning of each session, our agent uses these inputs to predict the sequence of all subgoals. Our key insight is that to better predict subgoals-to-do, it is also important to infer \textit{what has been done}. As such, we propose to additionally predict the completed subgoals, and include the agent's action history as an input to support the prediction. 

To take advantage of the power of pre-trained language models for this type of problem, all inputs and outputs are translated into language form. First, we convert the agent's action history into synthetic language (e.g. \texttt{PickUp(Cup)} $\rightarrow$ \textit{``get cup''}), and feed it together with the history of dialog utterances into the encoder. We then decode language expressions for subgoals one by one in an autoregressive manner.
As the raw outputs from the decoder can often be noisy due to language expression ambiguity or incompleteness, we add a \textit{natural-in, structure-out} decoder which learns to classify each of the subgoal components into its symbolic form,
and transform them to a language phrase as decoder input to predict the next subgoals.

\vspace{-5pt}
\subsection{Online Symbolic Planning}
\vspace{-5pt}
\label{sec:plan}



Symbolic planners excel at generating reliable and interpretable plans.
Given predicted subgoals and a constructed spatial-symbolic representation, PDDL \cite{McDermott1998PDDLthePD} planning algorithms can be applied  to generate a plan for each subgoal.\footnote{See Appendix \ref{sec:appendix_pddl} for more details on PDDL planning. } 
These short-horizon planning problems reduce the chance of drifting from the plan during execution. 
Nonetheless, failures are bound to happen during execution. 
A notable advantage of our approach is the transparency of its reasoning process, which not only allows us to examine the world representation and plan, but also gives the agent some awareness about potential exceptions and enables the development of mechanisms for replanning. 
In this section, we introduce several new mechanisms to make online symbolic planning feasible and robust in a dynamic physical world. 



\vspace{-5pt}
\paragraph{Finding Unobserved Objects} 
The agent's partial observability of the environment may cause a situation where in order to complete a subgoal, the agent needs an object that has not been observed yet. 
In this case, a traditional symbolic planner cannot propose a plan, and thus will fail the task. 
To circumvent this shortcoming, we extend the planner by letting the agent search for the missing object(s).
Specifically, during planning, our agent assumes that all objects relevant to subgoals exist, mocking full observability. If it has not observed any instance of an object required to satisfy the subgoal, then the agent plans with an assumed dummy object instance, which is assigned an additional state predicate \texttt{unobserved}. By incorporating these dummy objects, a plan can still be produced. For example, if the subgoal is \texttt{(Mug,isPickedUp)} but the agent has not observed a mug, the plan will become \texttt{Search(Mug), GoTo(Mug), PickUp(Mug)}. During execution, the agent is equipped with a dedicated \texttt{Search} action, which lets the agent explore the environment until an instance of the target object class has been observed and added to the world representation.\footnote{\addafterrebuttal{Details on object search can be found in Appendix~\ref{sec:appendix_map}.}}

\vspace{-5pt}
\paragraph{Search Space Pruning}
Large object and state search space slows symbolic planner down and creates unbearable latency to the system, as also found in \citet{pmlr-v164-agia22a}. 
We propose to solve this problem by pruning the search space to only include the relevant object instances. First, the planner finds all relevant object types for a subgoal predicate. For example, if the subgoal is to cook an object, then all object classes that can be used to cook (e.g., Microwave, Stove Burner, Toaster, etc.) are seen as relevant, even if the class is not explicitly stated in the subgoal.
Once relevant object classes are determined, the agent collects relevant instances of those classes. If there are other instances tied to the state of an instance, they are also deemed relevant. For example, if an object is inside of a receptacle, that receptacle object instance is also deemed relevant in the search space. All irrelevant instances are then discarded to speed up planning.

\begin{figure}[t!]
	\centering
    \includegraphics[width=0.97\columnwidth]{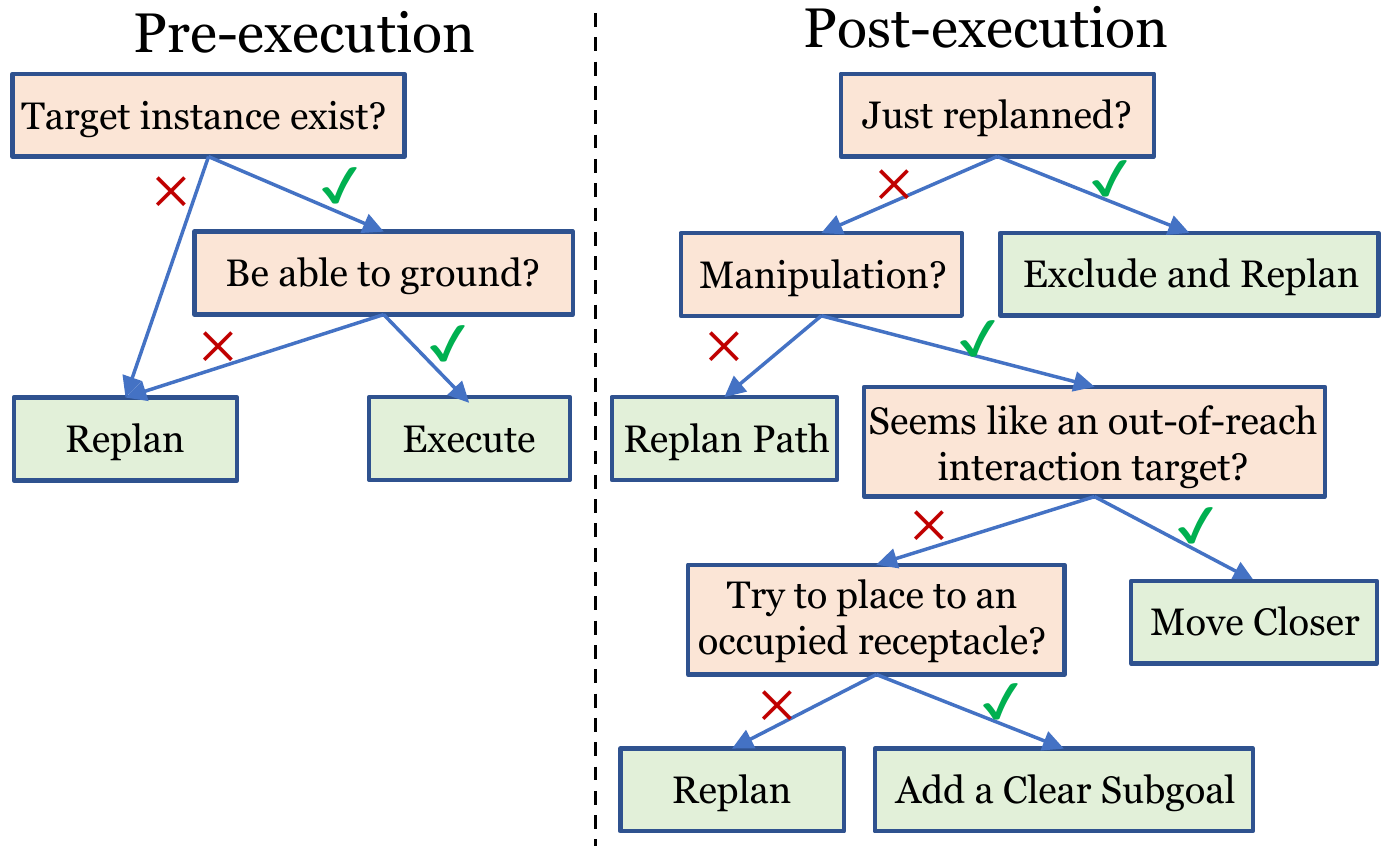}
    \vspace{-5pt}
    \caption{Decision trees for handling exceptions encountered during plan execution.  We term pre-execution errors as infeasible actions that cannot be instantiated in the situation, and post-execution errors as those when no outcome is observed after the action is executed. Red boxes denote situations based on the agent's belief. Green boxes denote exception handling strategies. }
	\label{fig:eh}
	\vspace{-12pt}
\end{figure}

\vspace{-5pt}
\paragraph{Action Failure Recovery}
The ability to detect exceptions is critical, as the agent's actions may fail for unexpected reasons.
For example, an agent may try to place an object in a receptacle, but the action fails because the receptacle is full. 
We can address this by applying exception-specific strategies; in this case, the agent should try to clear out the receptacle first.
Figure~\ref{fig:eh} shows a decision tree of recovery mechanisms that can be used to handle various types of these exceptions.
If recovery fails, this replanning process repeats until a step limit threshold is hit, at which point the agent gives up and moves on to the next subgoal.
While we manually define recovery strategies as an initial step, other ways to acquire or learn these strategies can be explored in the future.

\vspace{-5pt}
\section{Experiments And Result Analysis}

\vspace{-5pt}
\subsection{Experimental Setup}
We follow the evaluation methods originally proposed for TEACh, reporting metrics for task completion success and efficiency. First, \textit{task success rate} is the proportion of task sessions for which the agent met all goal conditions. \textit{Goal condition success rate} is more lenient, measuring the average proportion of goal conditions that were met in each session. Each metric has a path length weighted (PLW) counterpart, where the agent is penalized if it takes a longer sequence of actions than the human-annotated reference path. Given a metric $M$, the PLW version of $M$ is calculated by 
$M_{PLW} = \frac{\hat{L}}{max(\hat{L}, L_{*})} * M$,
where $\hat{L}$ and $L_{*}$ are the lengths of the predicted and reference path, respectively. 
There are also seen and unseen divisions for each split, where seen and unseen indicate whether the rooms in that division appear in the train set.


\begin{table*}[t!]
\centering
\resizebox{.99\linewidth}{!}{
\begin{tabular}{lcccccccc}
\toprule
\multirow{3}{*}{Model}&\multicolumn{2}{c}{Validation Seen} & \multicolumn{2}{c}{Validation Unseen}
&\multicolumn{2}{c}{Test Seen} & \multicolumn{2}{c}{Test Unseen}   \\ 
\cmidrule(lr){2-3} \cmidrule(lr){4-5} \cmidrule(lr){6-7} \cmidrule(lr){8-9} 
& Success & Goal-Cond & Success &Goal-Cond & Success & Goal-Cond & Success &Goal-Cond \\
\midrule
\textbf{\textit{Reactive}} &&&&&&& \\
ET       & 8.28 (1.13)  &  8.72 (3.82)   &  8.35 (0.86)  &  6.34 (3.69)   &  8.82 (0.29)   &  9.46 (3.03)   &  7.38 (0.97)  &  6.06 (3.17)  \\
HET      & 7.95 (1.18)  &  10.61 (5.73)  &  8.63 (1.77)  &  9.54 (5.59)   &  9.48 (0.85)   &  9.35 (4.94)   &  9.99 (1.71)  &  9.47 (5.11)  \\
HET-ON   & 12.58 (3.38) &  16.85 (10.21) &  12.52 (3.73) &  15.10 (9.53)  &  11.44 (3.47)  &  16.07 (9.69)  &  12.32 (3.12) &  14.81 (9.30) \\
\midrule
\textbf{\textit{Deliberative}} &&&&&&& \\
\acronym{}-PaP & 13.91 (8.15) &  21.74 (19.52) &  14.94 (6.03) &  19.80 (16.99) &  17.32 (7.43) &  19.14 (17.35) &  13.45 (6.20) &  18.82 (18.17)  \\
\acronym{} (Ours)     & \bf 16.89 (9.12) & \bf 25.10 (22.56) & \bf 17.25 (7.16) & \bf 23.88 (19.38) & \bf 18.63 (9.41) & \bf 24.77 (21.90) & \bf 16.71 (7.33) & \bf 23.00 (20.55)  \\
\bottomrule
\end{tabular}}
\vspace{-5pt}
\caption[Caption for LOF]{Task and Goal-Condition success rates on the TEACh EDH benchmark. The path-length-weighted version of metrics are in (parentheses). The highest values per column are in {\bf bold}. }
\vspace{-10pt}
\label{tb:benchmark} 
\end{table*}

\vspace{-5pt}
\paragraph{Implementations} 
Our sequence-to-sequence subgoal predictor is implemented by the BART-Large language model~\cite{lewis-etal-2020-bart}. 
For depth estimation, we fine-tune the depth model from \citet{blukis2022hlsm} on TEACh. For object instance detection, we use mask2former~\cite{cheng2022masked} as our panoptic segmentation model and fine-tune CLIP~\cite{openai_clip} to classify physical states.
Lastly, we use Fast Downward~\cite{Helmert2006FastDownward} for symbolic planning. More details can be found in the Appendix. 


\paragraph{Baselines} \hspace{-10pt} We compare \acronym{} to several baselines:

\begin{itemize}[leftmargin=*]
\itemsep0em
\vspace{-2pt}
    \item \textbf{Episodic Transformer (ET)}: End-to-end multimodal transformer model that encodes both dialog and visual observation histories, and predicts the next action and object argument based on the hidden state \cite{pashevich2021episodic}. 
    We reproduced the results based on the official code.\footnote{\url{https://github.com/alexpashevich/E.T.}} 
    \item \textbf{Hierarchical ET (HET)}:
    A hierarchical version of ET where low-level navigation actions are abstracted out by a navigation goal (e.g. \texttt{GoTo Mug}), then predicted by a separate transformer.
    \item \textbf{Oracle Navigator (HET-ON)}: HET with the navigator replaced by an oracle navigator that can obtain the ground truth target object location from the simulator's metadata.
    \item \textbf{\acronym{}-PaP}: A version of \acronym{} where symbolic planning is replaced by procedures as programs (PaP), i.e., manually defined rule-based policies following \citet{zhou2021hierarchical}.
    \item \addafterrebuttal{\textbf{JARVIS} \cite{zheng2022jarvis}: 
    A neuro-symbolic model similar to \acronym{} in spirit that also leverages a language model for subgoal planning and constructs maps for navigation. However, low-level plans for each subgoal are manually defined instead of being generated by a symbolic planner. }
\end{itemize} 

\vspace{-5pt}
\subsection{Overall System Performance}
We compare our agent with baseline models on the EDH task from TEACh in Table \ref{tb:benchmark}.\footnote{\addafterrebuttal{The results are based on the new validation/test split as described in the official TEACh repository (\url{https://github.com/alexa/teach\#teach-edh-offline-evaluation}), as the original test sets are not publicly available. Note that JARVIS is not listed here because their results are based on the original data split, which is not directly comparable to ours. A comparison with JARVIS on the original validation set is listed in Table \ref{tb:original_valid} in Appendix~\ref{apx: original split}.}}

\paragraph{Success Rate}
\acronym{} consistently outperforms all baseline models across both splits in every task and goal condition success rate metric. It outperforms HET-ON, the best reactive baseline equipped with an oracle navigator, by a sharp margin, demonstrating the advantage of our world representation and navigation planning.
It also outperforms the deliberative \acronym{}-PaP baseline, showing the value of employing existing mature automatic planning tools in the system rather than handcrafting procedural knowledge for the agent. 

\addafterrebuttal{
As shown in Table \ref{tb:per_task}, there is a high variance in the success rate of different task types, ranging from 10.6\% up to 57.4\%. The inverse correlation between the task success rate and the average number of task goal conditions indicates that longer-horizon tasks are more challenging, as expected. }


\begin{figure}[t]
	\centering
    \includegraphics[width=0.9\columnwidth]{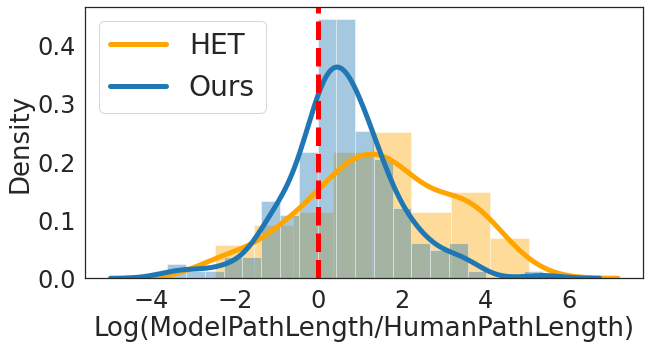}
    \vspace{-8pt}
    \caption{Comparison between HET and \acronym{} of relative success trajectory path lengths against ground-truth human trajectory lengths. The red dashed line denotes the same length as human trajectories, thus area to its left corresponds to the proportion of model predictions that achieved better efficiency than human performance.}
	\label{fig:traj_len}
	\vspace{-10pt}
\end{figure}

\vspace{-5pt}
\paragraph{Efficiency} 
For task completion efficiency, we find that \acronym{} again outperforms all baselines in PLW metrics, including HET-ON by over 10\% in PLW goal condition success rate.
We further compare the predicted trajectory lengths of our agent with the HET baseline and ground truth human data in Figure~\ref{fig:traj_len}. \acronym{} outperforms HET in efficiency by a large margin, with 26\% of tasks surpassing human efficiency. These results show the superiority of the deliberative agents in creating both accurate and efficient plans for task completion.

\begin{table}[t]
\centering
\resizebox{.95\linewidth}{!}{
\begin{tabular}{lcccc}
\toprule
\multirow{3}{*}{Task Name} & \multirow{3}{*}{\#Session} & \multirow{3}{*}{Avg. \#GCs}  & \multicolumn{2}{c}{\acronym{}}   \\ 
\cmidrule(lr){4-5}
&& & SR & GC \\
\midrule
WaterPlant & 61 & 1.0 & 57.4 & 59.3 \\
PutInOne & 95 & 1.9 & 26.3 & 38.1 \\
Boil & 82 & 2.2 & 19.5 & 29.8 \\
PutAll & 90 & 2.2  & 26.7 & 33.0 \\
Coffee & 122 & 2.8  & 27.9 & 41.9 \\
CleanAll & 114 & 2.8  & 12.3 & 27.3 \\
Toast & 201 & 7.1  & 15.4 & 27.8 \\
Slices & 200 & 7.1  & 15.5 & 20.8 \\
CookSlices & 243 & 7.5  & 11.9 & 30.2 \\
Breakfast & 380 & 8.7  & 14.7 & 21.7 \\
Sandwich & 287 & 9.0  & 13.9 & 22.9 \\
Salad & 274 & 10.9  & 10.6 & 15.8 \\
\bottomrule
\end{tabular}}
\caption{Per task performance on the unseen split. Tasks are sorted in a decreasing order w.r.t. the number of average goal conditions.}
\label{tb:per_task} 
\end{table}


\vspace{-5pt}
\subsection{Generalization to New Tasks}
\vspace{-3pt}
\addafterrebuttal{To demonstrate the generalizability of our approach, we additionally applied \acronym{} to the trajectory from dialog (TfD) subtask of TEACh without re-training any modules.
The major difference between EDH and TfD is the form of inputs given at each timestep. 
In EDH, the agent receives the contextualized interaction history (including dialog and physical actions) up to the current time, while in TfD, the agent is given the full session dialog without context, and must infer actions to complete the task induced by the dialog (analogous to instruction following).
As shown in Table \ref{tb:tfd}, \acronym{} significantly outperformed both the ET baseline and JARVIS, achieving success rates up to about 8\%, compared to a maximum of 1.8\% from baselines which are specifically trained on the TfD data.
}


\begin{table}[t]
\centering
\resizebox{.99\linewidth}{!}{
\begin{tabular}{lcccc}
\toprule
\multirow{3}{*}{Model}&\multicolumn{2}{c}{Valid Seen} & \multicolumn{2}{c}{Valid Unseen}   \\ 
\cmidrule(lr){2-3} \cmidrule(lr){4-5} 
& SR & GC & SR & GC \\
\midrule
ET                  & 1.02 (0.17) &  1.42 (4.82) &  0.48 (0.12) &  0.35 (0.59)  \\
JARVIS              & 1.70 (0.20) &  5.40 (4.50) &  1.80 (0.30) &  3.10 (1.60)  \\
\acronym{} (Ours)   & \bf 4.97 (1.86) & \bf 10.50 (10.27) & \bf 7.98 (3.20)  & \bf 6.79 (6.57) \\
\bottomrule
\end{tabular}}
\caption{\addafterrebuttal{Results on the TfD subtask of TEACh.}}
\vspace{-10pt}
\label{tb:tfd} 
\end{table}



\vspace{-5pt}
\subsection{Analysis of Submodules}\label{sec: analysis of submodules}
\begin{figure}[t!]
\centering
\includegraphics[width=0.97\columnwidth]{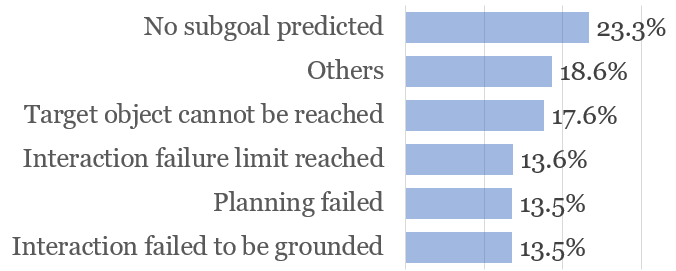}
\caption{Distribution of unrecoverable errors identified by our agent that cannot be recovered.}
\label{fig:unrecoverable}
\vspace{-10pt}
\end{figure}

\vspace{-3pt}
\addafterrebuttal{We perform further analysis to better understand the contribution and failures from each component.
When a session fails, the agent logs a possible reason based on its awareness of the situation. Figure \ref{fig:unrecoverable} shows the distribution of the errors.}

\addafterrebuttal{Subgoal prediction is a major bottleneck \addafterrebuttal{(accounting for 23.3\% of failures)}, since the agent only takes actions for subgoals that are predicted by this module. Most remaining errors are related to planning and exception handling. Inability to find a plan accounts for 13.5\% of failures, since the agent can only execute actions that it has planned. Meanwhile, 17.6\% of failures are caused by not finding an instance of a planned target object after searching for it. 
Interaction and grounding failures, where interaction with the virtual environment fails and the planner's exception handling mechanisms fail to resolve the issue, account for a respective 13.6\% and 13.5\% of errors.
Other types of failures are categorized as \textit{others}, which are typically due to inaccuracies in the agent's world representation.}

\addafterrebuttal{As most system failures are caused by the subgoal predictor and planner (including built-in exception handling techniques), we will discuss these modules' performance in terms of module-specific results (Tables \ref{tb:sg}-\ref{tb:planning}, Figure \ref{fig:exception_handling}), and their ablated contribution to the end task performance (Table \ref{tb:ablation}).}

\vspace{-3pt}
\subsubsection{Subgoal-Based Task Monitoring}
We conduct ablations on our subgoal-based task monitor from the perspective of subgoal prediction accuracy and overall task completion.
\vspace{-5pt}
\paragraph{Subgoal Prediction Accuracy}
\addafterrebuttal{In Table~\ref{tb:sg}, we examine the subgoal prediction accuracy of our subgoal prediction module (shown in Figure \ref{fig:subgoal_learning}).
First, we find conditioning on action history in addition to the dialog history achieves much higher prediction accuracy, suggesting this is a strong signal for the model to infer the progress and predict which subgoals still need to be completed in the task.
Further, we find that converting actions from unnatural symbols into language improves the prediction performance, showing the advantage of translating symbols to natural language inputs when using pre-trained language models. 
Finally, we show that conditioning on the model's predicted completed subgoals also improves the performance (as opposed to inferring them implicitly).}

\begin{table}[t!]
\centering
\resizebox{.99\linewidth}{!}{
\begin{tabular}{lccc}
\toprule
Model & Patient & Predicate & Destination   \\ 
\midrule
\acronym{}  & \textbf{81.1} & \textbf{65.6} & \textbf{76.4} \\
\midrule
\small ~~w/o Action History       & 71.4 & 60.1 & 69.5 \\
\small ~~w/o Actions in Language      & 73.6 & 57.4 & 74.2 \\ 
\small ~~w/o Pred. Completed SG   & 77.8 & 63.8 & 76.3 \\

\bottomrule
\end{tabular}}
\vspace{-5pt}
\caption{\addafterrebuttal{Subgoal prediction accuracy of \acronym{} on the validation unseen split, compared to ablations that do not condition on the action history, language forms of the actions, and self-predicted completed subgoals.}}
\vspace{-5pt}
\label{tb:sg} 
\end{table}

\begin{table}[t]
\centering
\resizebox{.99\linewidth}{!}{
\begin{tabular}{lcccc}
\toprule
\multirow{3}{*}{Model}&\multicolumn{2}{c}{Valid Seen} & \multicolumn{2}{c}{Valid Unseen}   \\ 
\cmidrule(lr){2-3} \cmidrule(lr){4-5} 
& SR & GC & SR & GC \\
\midrule
\acronym{}                        & \textbf{16.9} (9.1) &  \bf 25.1 (22.6) & \bf 17.3 (7.2) &  \bf 23.9 (19.4)  \\
\midrule
\small ~ last subgoal only      & \textbf{16.9} (\textbf{12.2}) &  19.0 (21.4) &  12.4 (5.6) &  12.9 (14.6)   \\
\small ~ w/o scene pruning      & 14.9 (8.2)  &  20.2 (20.3) &  11.8 (5.9) &  14.2 (15.6)  \\
\small ~ w/o replanning         & 14.9 (8.2)  &  22.5 (20.4) &  14.9 (6.9) &  19.0 (17.5)  \\
\bottomrule
\end{tabular}}
\caption{Ablation study of components with respect to the end task performance. \textit{Last subgoal only} refers to directly working on the last predicted subgoal instead of all predicted subgoals one after another. \textit{Without scene pruning} refers to using the planner without search space pruning. \textit{Without replanning} refers to disabling replanning from the action failure recovery strategy. }
\vspace{-10pt}
\label{tb:ablation} 
\end{table}

\vspace{-5pt}
\paragraph{Task Completion}
In the second row of Table~\ref{tb:ablation}, we see the impact of monitoring all subgoals rather than the current one on the end task performance. In 6 out of 8 metrics, \acronym{}, which learns to predict all subgoals throughout the task session, achieves a significant improvement. 
A possible explanation of the improvement is that some subgoals can only be addressed when a previous one is completed,\footnote{For example, \texttt{(BreadSlice, isPlacedTo, Plate)} requires \texttt{(BreadSlice, isSliced}) to be achieved first for bread slices to exist in the domain.} while not all subgoals are strictly conditioned on previous ones.\footnote{For example, \texttt{isSliced(Bread)} entails addressing \texttt{isPickedUp(Bread)} along the way.}

   

\vspace{-3pt}
\subsubsection{Online Symbolic Planning}
We also conduct ablations for our online planner, specifically in its capabilities for search space pruning and action failure recovery. 

\vspace{-3pt}
\paragraph{Finding Unobserved Objects}
\addafterrebuttal{The planner depends heavily on the constructed spatial-symbolic map, and errors here may cause errors in scene navigation, especially when target objects are hidden in the environment.
As 17.6\% of failures are caused by not finding the target object, we made a further analysis on our agent's object search performance. We found that \acronym{} performs at least one object search action in 11.8\% of subgoals across 21.7\% of the game sessions. Among these sessions, the success rate is 3.7\%, which is far below the overall success rate of 17.3\%. This indicates object search is indeed a big performance bottleneck. To improve performance, future work may develop stronger methods to locate missing objects.}

\vspace{-3pt}
\paragraph{Search Space Pruning}
\label{sec:pruning}
As seen in Table \ref{tb:planning}, scene pruning sharply improves both planning success (i.e., the rate at which formed PDDL problems had a solution generated for them) and the average runtime of the planner, which greatly speeds up online execution. Table \ref{tb:ablation} also depicts task and goal-condition success rates for our model without scene pruning, showing that scene pruning contributes to \acronym{}'s success, and that improving PDDL plan search complexity improves overall task success. 

\begin{table}[t]
\centering
\resizebox{.99\linewidth}{!}{
\begin{tabular}{ccc}
\toprule
Method & Plan Search Fail Rate & Avg. Runtime \\
\midrule
w/o scene pruning & 24.3\%  & 94.64s \\
w/ scene pruning  & \bf 9.1\%  & \bf 4.37s \\
\bottomrule
\end{tabular}}
\caption{PDDL planning success rate and average runtime. We set the overall time-out for domain translation and solution searching to be 120s in our experiments. 
}
\vspace{-5pt}
\label{tb:planning} 
\end{table}

\begin{figure}[t!]
	\centering
    \includegraphics[width=0.99\columnwidth]{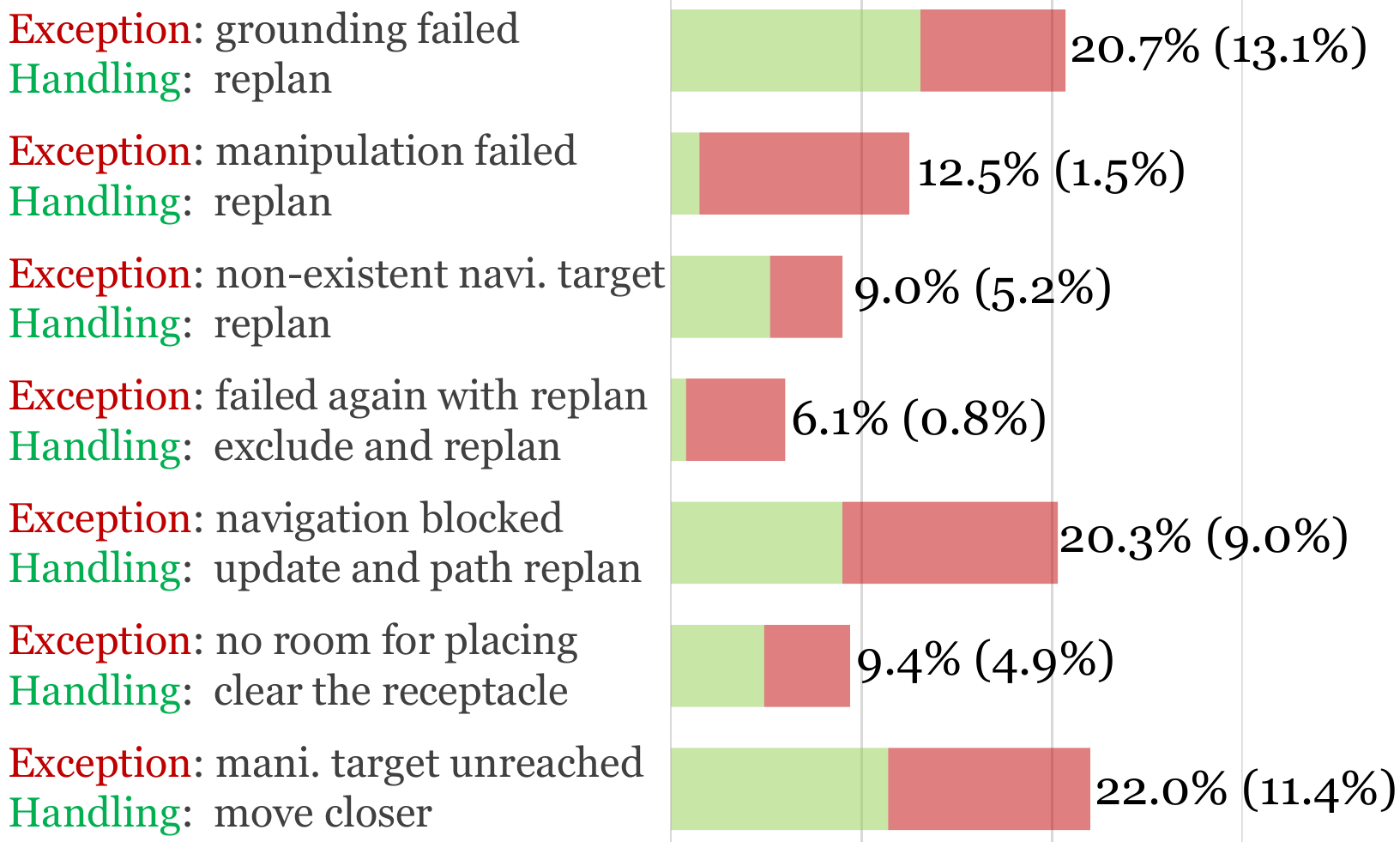}
    \caption{Distribution of \textcolor{Maroon}{execution exceptions} encountered and \textcolor{ForestGreen}{self-handled} by our agent. The proportions of self-handled exceptions are in parentheses. The results are based on 2708 exceptions encountered in 1078 sessions from the validation unseen split.}
	\label{fig:exception_handling}
	\vspace{-10pt}
\end{figure}


\vspace{-5pt}
\paragraph{Action Failure Recovery}
\addafterrebuttal{\acronym{}'s exception handling mechanisms enable it to identify and recover from exceptions to prevent some types of potential task failures. These
mechanisms work in symbolic space, and are fully interpretable.
In Figure \ref{fig:exception_handling}, we show the distribution of exception types the agent encountered and how it attempted to recover.}
A major source of exceptions is the mismatch between the agent's perceived internal world state to the real situation. This is because the internal state is updated per timestep, while a generated plan remains unchanged unless it is not applicable anymore. Typical cases include failure to ground to the interaction target, being blocked by an obstacle due to the change of estimated destination, or attempting to navigate to an object that no longer exists. 
We find that replanning can handle about half of these failure cases. Within these cases, our proposed mechanisms for addressing exceptions with receptacle occupancy and object reachability contribute to around 15\% of recoveries.
Meanwhile, when a manipulation fails (i.e., no effect observed) and replanning does not help, our agent can only handle about 12\% of these cases. 
Future work can apply further analysis and exception handling techniques to mitigate this type of failure. 
\addafterrebuttal{As shown in the last line of Table \ref{tb:ablation}, replanning contributes up to 2.4\% of the overall task success rate.} 
\section{Related Work}
\vspace{-5pt}


The topics most relevant to this work include: semantic scene representation, hierarchical policies, and symbolic planning.
\paragraph{Semantic Scene Representation} A strong world representation is necessary for spatial and task-structure reasoning in embodied AI. Prior work has explored using vectorized representation \cite{pashevich2021episodic,Nguyen2021LookWA,Suglia2021EmbodiedBA,Khandelwal2021SimpleBE}, 3D scene graphs~\cite{3dsg2019, 3dsg2020, pmlr-v164-li22e, pmlr-v164-agia22a} and semantic map \cite{chaplot2020object, saha2021modular, blukis2022hlsm, min2021film}.
We follow the latter approach. But unlike prior work, our semantic map additionally models object \textit{instance} and \textit{physical state} to inform planning, and enables the use of off-the-shelf symbolic planners.

\paragraph{Hierarchical Policies} 

Hierarchical policy learning and temporal abstraction have long been studied in reinforcement learning \cite{hier_old2, hier_old3,hier_old4,hier_old5,hier_old6,hier_old7,hier_old8} to uncover the inherent structures of complex tasks. Recently, such hierarchical approaches are gaining interest in natural language instruction following \cite{sharma2021skill, zhang2021hitut, zhou2021hierarchical, blukis2022hlsm,zheng2022jarvis}. 
While prior work applies planning at the high level for subgoal prediction, a reactive approach is still used for low-level action execution. As such, we build a deeply deliberative agent by using specialized planners for interaction and navigation actions based on our spatial-symbolic world representation, thus enabling more reliable subgoal completion. 
\paragraph{Symbolic Planning}
Symbolic planning has been extensively studied in classical AI literature and applied in robotics, where  
entities, actions, and goals are represented as discrete symbols \cite{McDermott1998PDDLthePD,galindo2004improv, kavraki1996prm, galindo2008semmap}.
Recent work aims to learn these symbols from language \cite{she2017interactive, zellers2021piglet, silver2021learn, MacGlashan2015GroundingEC} and vision \cite{lamanna2021online,xu2022sgl}.
Unlike these efforts, our approach combines both modalities by learning symbolic task goals from natural language, and leveraging a persistent scene representation based on visual object detection. 
Further, unlike past work, we optimize our approach for embodied instruction following by adding capabilities to search for missing objects, knowledgeably prune the search space, and recover from failed actions.

\vspace{-5pt}
\section{Conclusion \addafterrebuttal{and Discussion}}
\vspace{-5pt}
\label{sec:discussion}
This paper introduced \acronym, a deliberative instruction following agent which outperformed reactive agents \addafterrebuttal{in terms of both success rate and efficiency} on a challenging benchmark TEACh. 
\addafterrebuttal{The use of explicit symbolic representations also supports a better understanding of the problem space as well as the capabilities of the agent. }

\addafterrebuttal{Despite the strong results, we are still far away from building embodied agents capable of following language instructions, and we identify three major bottlenecks that will motivate our future work. First, the capability to \textit{handle uncertainty in perception and grounding of language} is essential when constructing symbolic representations for planning and reasoning. We hope to explore tighter neuro-symbolic integration in order to allow better learning of symbolic representations, while still keeping the advantage of providing transparent reasoning and decision-making in agent behaviors. Second, the capability to \textit{robustly handle exceptions} encountered during execution is vital. Humans proactively seek help if they get stuck in completing a task, while current AI agents lack this capability. Integrating dialog-powered exception handling policies is our next step to extend \acronym{}. Third, future work needs to improve the \textit{scalability of symbolic representations} for use in novel situations. In this work, the symbol set for predicates and subgoals is predefined, which requires engineering effort and may not be directly applicable to other domains. As recent years have seen an exciting application of large language models for acquiring action plans \cite{ahn2022can}, we will build upon these approaches for more flexible learning of action planning in an open-world setting in the future.  }



 \section*{Limitations}
As in many embodied agents which apply planning algorithms, \acronym{} operates in a closed domain of objects and actions, which requires manually specifying object affordances as well as state changes caused by actions. This means that if faced with a task involving new objects and/or predicates, the agent will fail without updating the knowledge base and re-training perception modules.
Moreover, the agent is tied to its action space in the Thor simulator, and cannot easily acquire new actions. Future work can address this by exploring the possibility to acquire new objects and actions in this space, including object affordances and action effects. In addition, as most of the work currently is done in the simulated environment, whether the learned models can be successfully transferred to the physical environment remains a big question. Sim2real transfer has not been well explored in this line of work on embodied AI.

Furthermore, the evaluation applied to \acronym{} is completely automatic and only measures task completion information. We do not evaluate the safety of the agent's movements, or the disruption of its actions to the environment. This evaluation cannot capture a human user's trust and confidence toward the agent's behavior. Before such agents can be brought from the simulator space into the real world, extensive study will be required to ensure the safety and respect of human users.





\section*{Ethics Statement}


One main goal of the embodied agents in the simulated environment is to enable physical agents (such as robots) with similar abilities in the future. As such, one key ethical concern is the safety of those physical agents, e.g., the need to minimize the harm and damage they may cause to humans and their surroundings. The current setup in the simulated environment is not designed to address safety issues, mostly based on task success. In the future, we hope to incorporate such safety considerations into the design of intelligent agents, by means such as minimizing the occurrence of collisions, performing more consistent and predictable actions, and explicitly modeling human behaviors to avoid harm. We also hope to see more datasets and benchmarks emphasizing safety in their evaluation.

\section*{Acknowledgements}
This work was supported in part by an Amazon
Alexa Prize Award, NSF IIS-1949634, NSF SES-2128623, and DARPA PTG program HR00112220003. 
The authors would like to thank
the anonymous reviewers for their valuable comments and suggestions.

\bibliography{custom}
\bibliographystyle{acl_natbib}

\appendix

\section{Appendix}
\label{sec:appendix}


\subsection{TEACh vs ALFRED}
The ALFRED \cite{shridhar2020alfred} dataset is a feasible alternative to evaluate our method on, but we do not conduct our evaluations on it in this work. Here we provide some comparisons between the TEACh and ALFRED datasets to suggest that evaluating on TEACh is sufficient, due to its higher complexity and difficulty, and because this dataset necessitates a deliberative approach.
\begin{enumerate}
\item TEACh has many more state variations. For example, the sink in an ALFRED environment is always clear so the agent can always rinse an object by placing it into the sink and toggling the faucet on. In TEACh the sink might be occupied so the agent may have to clear up the sink first. Similarly, mugs are always empty and clean (if it is not the target object of a cleaning mug task) in ALFRED, while in TEACh they can be filled with water or dirty. The more physical state variations suggest that the TEACh dataset poses more challenges in understanding the situation.
\item In TEACh the agent has to learn multiple ways to fulfill one subgoal. As an example, in ALFRED, the cooking task corresponds to a rigid protocol of operating the microwave. In TEACh, there are multiple ways of manipulating different kitchen utilities to cook, such as either using a microwave or a pan to cook a potato slice, using a toaster to make toast, or boiling potatoes with either a pot and stove or a bowl and microwave. This also demonstrates TEACh's sufficiency in terms of planning complexity.
\item It is a known issue that solving a dataset synthesized by an oracle planner may result in reverse engineering of the data generation process. As TEACh consists of human-human interaction data, there is no concern that there exists reverse engineering.
\item To monitor task progress in TEACh, the agent has to infer what has been done before it can predict future subgoals. In ALFRED, all language instructions for the episode are provided at the beginning, so there is no need to estimate progress. In particular, a reactive agent suffers less in the less challenging ALFRED because it does not need to monitor task progress and reason about its plan. A deliberative agent, on the other hand, shines in TEACh.
\end{enumerate}

\begin{figure*}[t!]
    \centering
    \includegraphics[width=0.99\textwidth]{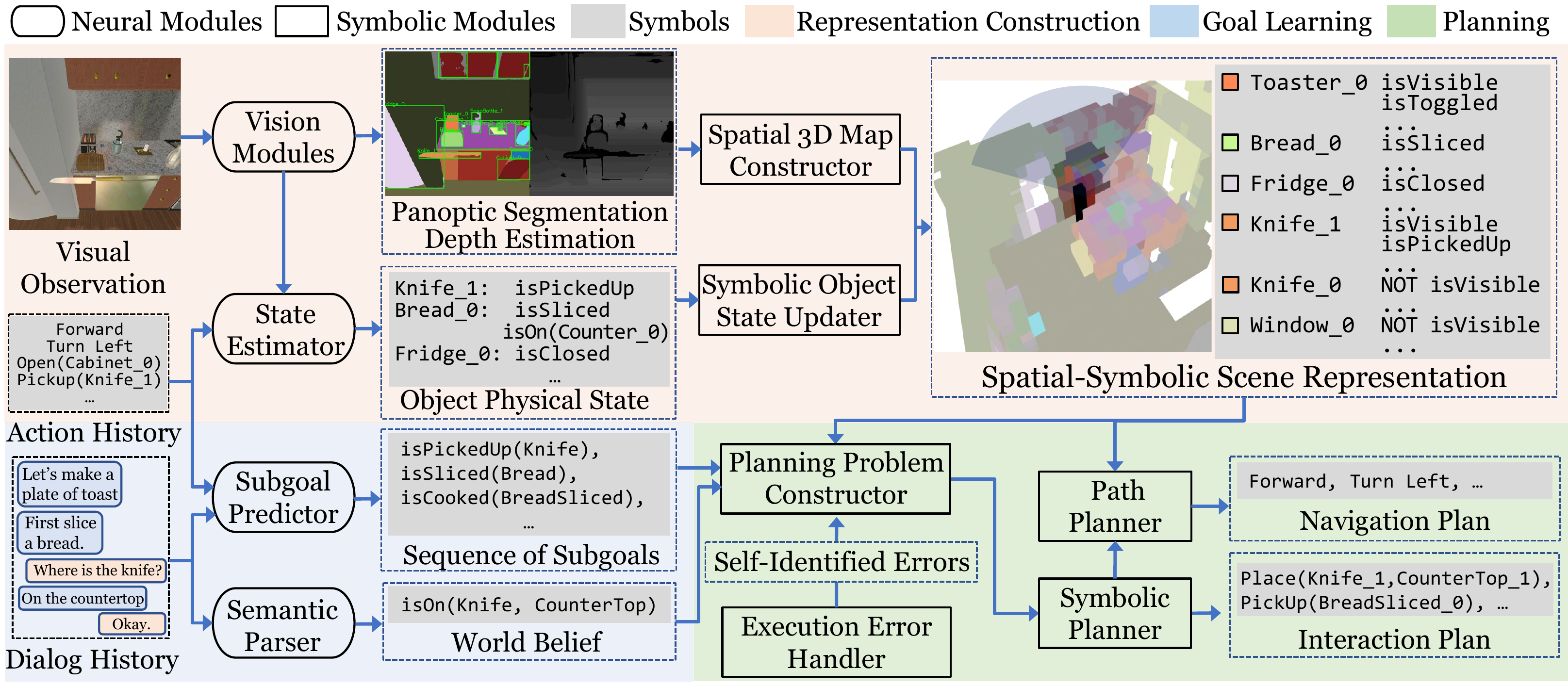}
    \caption{Overview of our neuro-symbolic system, which learns a rich, visually-informed semantic representation of the surrounding environment, and captures hierarchical tasks in two phases. First, it uses a sequence-to-sequence model to predict the sequence of high-level subgoals, both completed and upcoming, and additionally extracts some spatial relations from the dialog to supplement the world representation. For each subgoal, we apply a specialized planner (either for navigation or manipulation) over the world representation to accurately execute the low-level actions and achieve the subgoal.
    Best viewed in color.}
    \label{fig:overview}
\end{figure*}

\subsection{System Overview}

A detailed overview of all system components and sub-components is provided in Figure~\ref{fig:overview}.

\subsection{Perception System Details}

Our perception system consists of a panoptic segmentation model, a depth estimation model and a physical state estimation model. At each time $t$, after receiving the RGB image $I_t$ from the egocentric vision, our perception system predicts egocentric depth estimation $I_t^D$, panoptic segmentation $I_t^P$, and corresponding state estimation $I_t^S$ for each predicted objects in $I_t^P$. The output of this system serves as the foundation for building the symbolic representation for the agent, 

We use main-stream pretrained models for the standard depth estimation and panoptic segmentation task and implement a custom Vision-Transformer \cite{dosovitskiy2020vit} based model for physical state estimation. A dedicated dataset, TEACh-Perception, is collected for training these models.

\subsubsection{Data Collection}

We collect the visual perception data by replaying the TEACh dataset with additional filtering and data augmentation. Just like in the main experiment, we use the train split to build our training set, validation seen and validation unseen splits for two test sets.

We replay each of the games within the split and record egocentric images frame by frame, with their corresponding depth data, panoptic segmentation and object physical states ground truth extracted from the rendering engine. We find consecutive frames within the game can be very similar, degrading the dataset quality tremendously. Therefore, we propose to filter the frames on the fly. Namely, we use ImageHash \footnote{\url{https://github.com/JohannesBuchner/imagehash}} to calculate the similarity between the consecutive frames and only record the frame if its hash differs from the previous recorded frame by at least 3. To further improve dataset quality, we call the simulator's built-in lighting randomization and material randomization methods if the consecutive frames are similar by the aforementioned criteria. Following this method, we collect 179,144 annotated frames for the training set, 22,169 and 71,452 frames for the validation seen and unseen set respectively.

\subsubsection{Model Design \& Training}

\paragraph{Depth Estimation}
We follow the customized U-Net \cite{ronneberger2015u} depth estimation model introduced in \cite{blukis2022hlsm} and used its weight to initialize our model. To save computation, before passing $I_t$ into the U-Net, we first downsample $I_t$'s resolution from 900x900 to 300x300 and upsample the model's output back to 900x900 in the end. The model was finetuned with the same hyper-parameter  for 30,000 steps on a single Nvidia A40 GPU on our TEACh-Perception dataset.

\paragraph{Panoptic Segmentation}
Mask2Former \cite{cheng2022masked} is used as the panoptic segmentation model. We use the Swin-B 21K model \footnote{\url{https://github.com/facebookresearch/Mask2Former/blob/main/configs/cityscapes/semantic-segmentation/swin/maskformer2_swin_base_IN21k_384_bs16_90k.yaml}} in MMDetection \cite{mmdetection} for initialization, decrease the learning rate to $3 \times 10^{-5}$ and finetune it for 80,000 steps on 8 Nvidia A40 GPUs.

\paragraph{Physical State Estimation}
As each object has different affordances,\footnote{Object affordance is pre-configured in Thor~\cite{kolve2017ai2thor}, the virtual environment used to create TEACh.} we only predict the possible valid physical states for each object. For example, we will only predict a physical state value for the attribute \texttt{isFilledWithWater} for objects that are \texttt{Fillable}, e.g., bowls and cups.

We implement a custom model for state estimation tasks. The input consists of an object-centric image with its object's class id. After the image passes through an image encoder and class id through an embedding layer, two features are concatenated and go through two fully-connected layers to produce its state estimation. We use ViT-B/32 \cite{dosovitskiy2020vit} with an additional projection layer as the image encoder and initialize its weights from CLiP \cite{radford2021learning}. The projection layer is a linear layer followed by a ReLU function, projecting the 512-dimension image feature down to 128 dimensions. The class id embedding has a dimension of 8. The following 2 fully-connected layers are with ReLU activation function, each with output dimensions 128 and 3. The model is then trained for 150,000 steps on our TEACh-Perception dataset with a learning rate of $1 \times 10^{-5}$.

Lastly, our system also requires predicting objects' spatial relationship of whether or not one object is on another object. We use a simple rule-based algorithm to predict these spatial relations, which is, given object $A$ in class $C_A$ with bounding box $B_A = (x_{A1}, y_{A1}, x_{A2}, y_{A2})$, object $B$ in class $C_B$ with bounding box $B_B = (x_{B1}, y_{B1}, x_{B2}, y_{B2})$, where bounding boxes are in absolute coordinates notation, the model predicts $A$ is on $B$ if and only if 1.) from the prior receptibility knowledge, it is possible for $C_A$ to be on $C_B$, 2.) $(x_{B1} \leq (x_{A1} + x_{A2})/2 \leq x_{B2})\land(y_{A2} \geq y_{B1})\land( (y_{A1} + y_{A2})/2  \leq y_{B2})$.

During experimentation, we find three factors constraining state estimation accuracy and provide solutions respectively. 
First, the object-centric images are extracted from $I_t$ by objects' bounding boxes, which include little information about its surrounding. For example, to predict the \texttt{isToggled} state of a cabinet, the spatial relationship between the cabinet and the desk must be included in the image. To solve this problem, we enlarge the bounding box by 30\% on each side to include more spatial information. 
Secondly, we find when in distance, some objects will be too small to extract meaningful state information, hurting the accuracy of physical state estimation. Therefore, we filter out objects whose corresponding bounding boxes are less than 1000 pixels large in the egocentric image before passing into this model. 
Lastly, although whether or not a \texttt{StoveBurner} is turned on is mostly reflected on the existence of fire flames above it, the corresponding \texttt{isToggled} state is stored in the \texttt{Knob} object controlling the \texttt{StoveBurner}. To solve this misalignment between the visual feature and the corresponding state, we manually move the \texttt{isToggled} state from the knob to the corresponding \texttt{StoveBurner}.

\subsubsection{Results}
We additionally decouple our panoptic segmentation model and physical state estimation model from the whole system and evaluate them separately.  Note that we do not perform a dedicated evaluation for the depth estimation model as it is directly adopted from \cite{blukis2022hlsm} with minimal change.

\paragraph{Panoptic Segmentation}
In Table \ref{tb:pan-result}, we report the panoptic segmentation performance of our model on both validation seen and unseen splits. The significant performance drop on PQ of about 20\% when transferring from seen to unseen room setup indicates the difficulty in building a more generalized model, which we leave future works to explore.

\begin{table}[t]
\centering
\resizebox{.99\linewidth}{!}{
\begin{tabular}{lcccccc}
\toprule
\multirow{3}{*}{Object Type}&\multicolumn{3}{c}{Valid Seen} & \multicolumn{3}{c}{Valid Unseen}   \\ 
\cmidrule(lr){2-4} \cmidrule(lr){5-7} 
& PQ & SQ & RQ & PQ & SQ & RQ \\
\midrule
All & 70.65 & 84.68 & 80.17 & 52.91 & 71.90 & 60.25\\
\midrule
\small ~ Things & 71.36 & 85.11 & 81.40 & 56.37 & 75.10 & 64.37 \\
\small ~ Stuff & 65.43 & 81.77 & 71.71 & 32.95 & 53.46 & 36.49 \\
\bottomrule
\end{tabular}}
\caption{Performance of our panoptic segmentation model on valid seen and unseen split. PQ, SQ, RQ are panoptic quality metric, segmentation quality metric and recognition quality metric originally defined in \cite{kirillov2019panoptic}.}
\label{tb:pan-result} 
\end{table}

\paragraph{Physical State Estimation}
In Table \ref{tb:state-est}, we report the performance of our model on both validation seen and unseen splits. It is worth noting that although our model can perform well on \texttt{isDirty} state with an accuracy of over 95\%, \texttt{isToggled} state only has an accuracy of 77.0 \% and 74.7\%. After sampling and inspecting data points and the model's corresponding predictions, we argue that the performance gap between different states mainly comes from the difference in the states' visual saliency. For example, while \texttt{isDirty} state affects almost every part of the object, the effect of \texttt{isToggled} is much more inconspicuous, e.g. \texttt{isToggled} only affects the color of a small LED light on the \texttt{CoffeeMachine}.

\begin{table}[t]
\centering
\resizebox{.99\linewidth}{!}{
\begin{tabular}{lcc}
\toprule
\multirow{3}{*}{State Name}&{Valid Seen} & {Valid Unseen}   \\ 
\cmidrule(lr){2-2} \cmidrule(lr){3-3} 
& Accuracy & Accuracy \\
\midrule
All & 85.5 &  85.6 \\
\midrule
\small ~ \texttt{isDirty} & 95.8 & 96.5 \\
\small ~ \texttt{isFilledWithWater} & 82.7 & 86.2 \\
\small ~ \texttt{isToggled} & 77.0  &  74.7 \\
\bottomrule
\end{tabular}}
\caption{Accuracy of our state estimation model on both validation seen and unseen split. \texttt{isDirty}, \texttt{isFilledWithWater} and \texttt{isToggled} are the physical states that our system needs.}
\label{tb:state-est} 
\end{table}

\subsection{Map Construction and Navigation}
\label{sec:appendix_map}
\begin{figure*}[t!]
    \centering
    \includegraphics[width=0.99\textwidth]{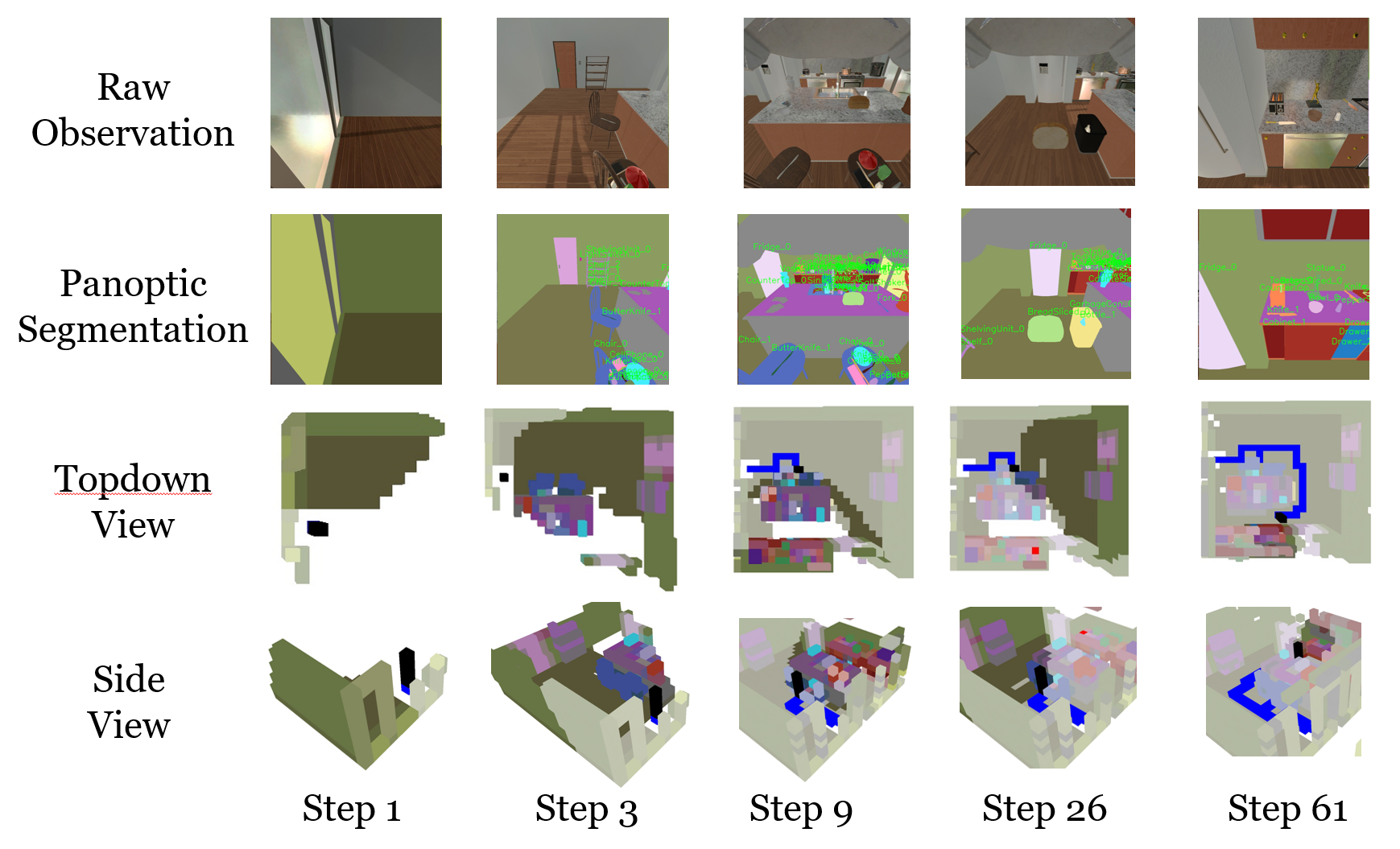}
    \caption{Illustration of the 3D voxel map construction and interaction process of a making toast task. The agent is labeled in black, and the agent path is labeled in blue. Current navigation target in the agent's plan is labeled in red. }
    \label{fig:path}
\end{figure*}

We describe below the transformation procedure of converting egocentric visual observations to a 3D semantic map.
Inspired by \citet{blukis2022hlsm}, we project the pixels from the panoptic segmentation model, with the help of a depth estimation model, to a 3D point cloud. The points in the point cloud are then binned into $0.25$m$^3$ cubes, called \emph{voxels}. Note that throughout this conversion process, object class information is preserved. The resulting map is $V_{t} \in [0, 1] ^{C \times X \times Y \times Z}$, where $C$ is the number of object classes and $X,Y,Z$ are the length, width and height of the map, respectively.
Given this 3D semantic voxel map, we can plan navigation paths using a deterministic path planner like the Value Iteration Network (VIN) introduced in \citet{blukis2022hlsm}. The path produced can be further converted to navigation action sequences, which can be directly followed by the agent through primitives. See Figure \ref{fig:path} for an example of map construction and path planning results in a single episode. \addafterrebuttal{When exploring for missing objects, the agent randomly samples an object from the 5 farthest objects in its current world representation and navigates to it. There is no specific timeout for the search action, but each subgoal is limited to a maximum of 200 action steps. }

\subsection{Instance Match Algorithm}
\label{sec:appendix_instance_match}
We use the following algorithm to match object instances from the current observation to instances already existing in the symbolic instance lookup table. If it is the first time that an instance is observed, we will register it into the lookup table. First, we group all the detections in the current observation based on its object type, and denote $n_c^{2D}$ as the count of object of type $c$. Then for each object type, we get object instances of the same type from the lookup table that is visible (at least one voxel falls into the visibility mask) at the moment, and denote the count as $n_c^{3D}$. If $n_c^{2D}==n_c^{3D}$, we match each pair of them to minimize the pair-wise distance computed as the Euclidean distance between their centroids. If $n_c^{2D}>n_c^{3D}$, we match up to $n_c^{3D}$ detections to the existing instances, and then register new instances into the lookup table. If $n_c^{2D}<n_c^{3D}$, we will match up to $n_c^{2D}$ detections to the existing instances, and remove the remaining instances. When a detection is matched to a instance in the lookup table, we update its voxel mask by performing a logical-or operation of the projected voxel mask of the detection and the existing voxel mask. Thus different partial observations of the same instance at different positions can get merged.

\subsection{Subgoal Prediction}
\label{app:subgoal}

\subsubsection{Subgoal Labeling}
To annotate the key subgoals for each task type in TEACh, we define a minimal set of object states that is essential for the completion of the task.
For example, as previously shown in Figure~\ref{fig:new overview}, \texttt{(Knife,isPickedUp)} and \texttt{(Bread,isSliced)} are subgoals for making toast, while \texttt{(Knife,isPlacedTo,Counter)} is not. This is because while placing a knife on the counter may often happen when making toast in the TEACh data, it is not essential to achieve the goal conditions for making toast. We then use an automatic subgoal labeling algorithm to get these subgoal annotations for the entire dataset.

\subsubsection{Subgoal Annotation}
\begin{figure}[t!]
    \centering
    \includegraphics[width=0.99\columnwidth]{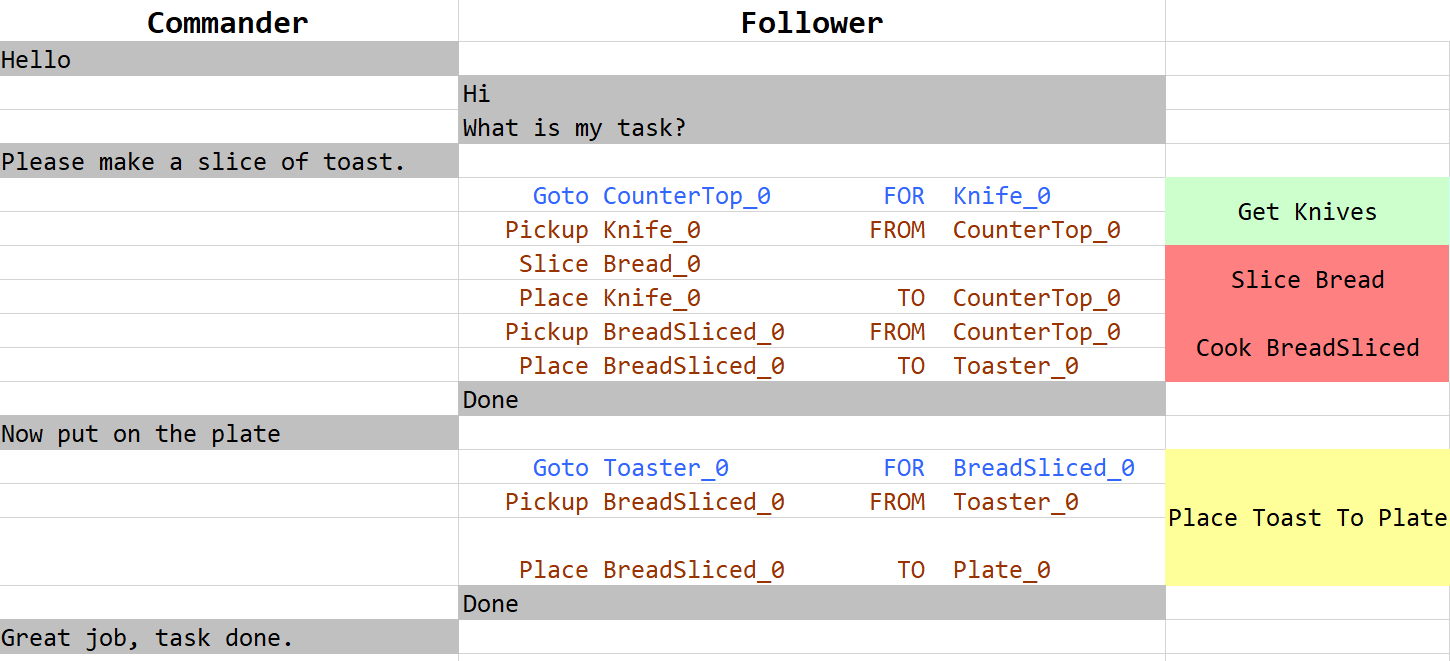}
    \caption{An example of subgoal annotations. }
    \label{fig:sg_ann}
\end{figure}

We propose a method to automatically annotate subgoals from the trajectories. Based on the metadata of task goal conditions and the oracle world state change of each step recorded in the dataset, we extract whether a goal condition is achieved at every step. If there is a goal condition completed, it is annotated as a subgoal being achieved in that step. We show an annotation example in Figure \ref{fig:sg_ann}.

\begin{figure*}[t!]
    \centering
    \includegraphics[width=0.99\textwidth]{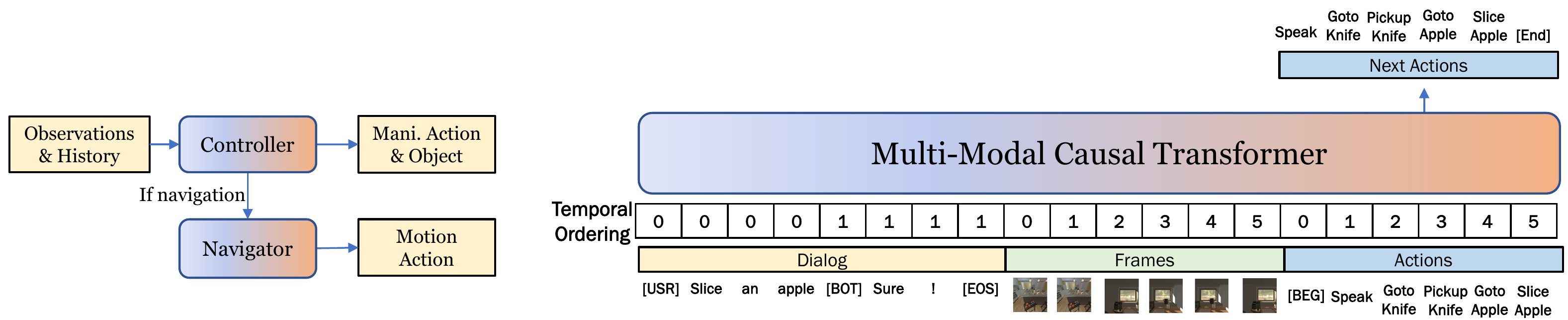}
    \caption{Architecture of the hierarchical ET model. }
    \label{fig:het}
\end{figure*}

\begin{figure*}[t!]
    \centering
    \includegraphics[width=0.99\textwidth]{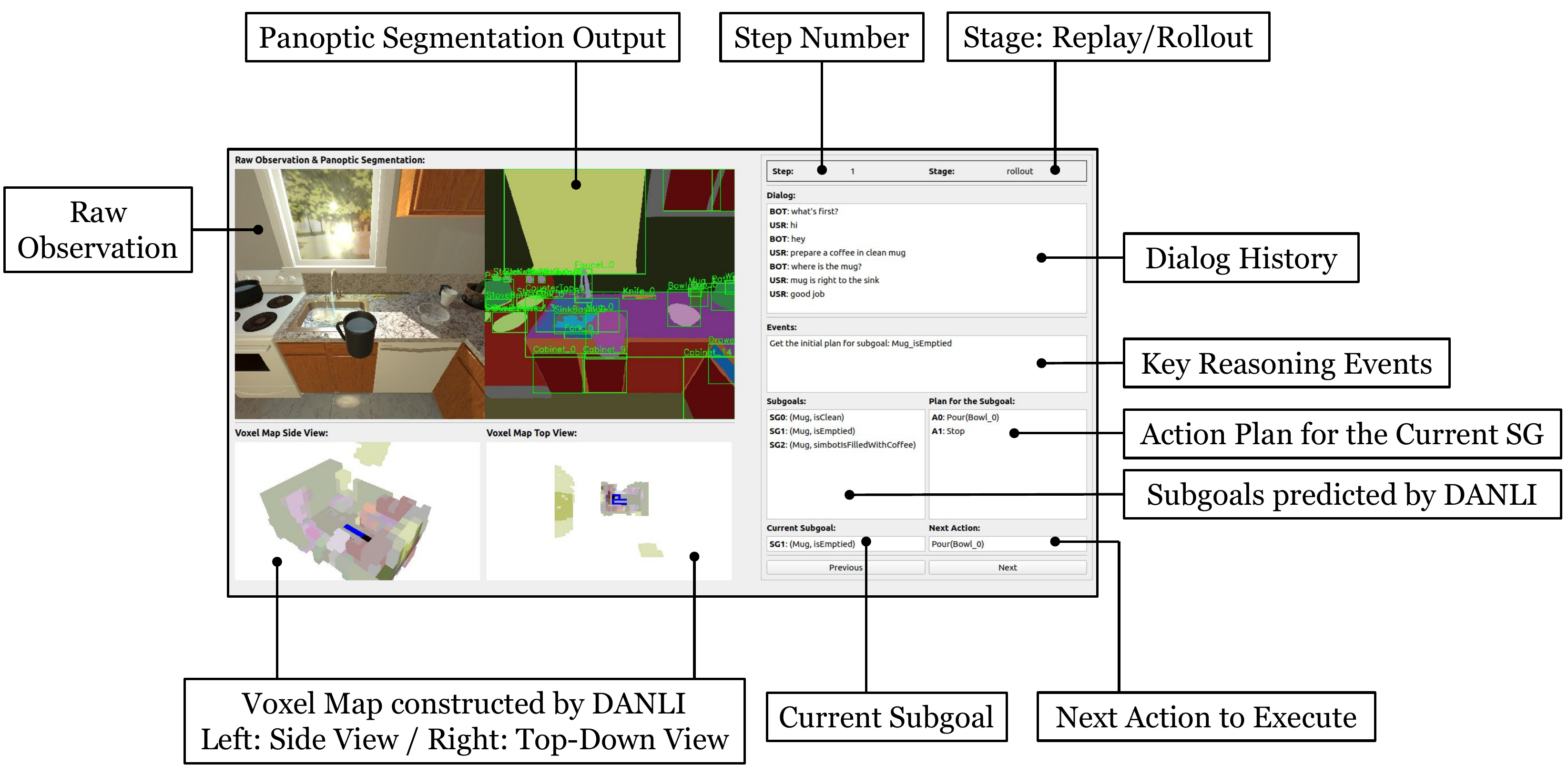}
    \caption{Demo of the Graphical User Interface (GUI) created for visualizing \acronym{}'s internal representations.}
    \label{fig:visualizer_demo}
\end{figure*}

\subsection{Symbolic Planning Details}
In this section we describe details of our symbolic planning module, including how PDDL planning works, the types of subgoal predicates we used in our experiments, and pseudocode to describe how our online planning pipeline works.

\subsubsection{PDDL Planning}\label{sec:appendix_pddl}
PDDL planners, such as FastDownward \cite{Helmert2006FastDownward} and ENHSP \cite{ENHSP}, require as inputs \textit{problems}, which symbolically specify the existing preconditions and desired goal, and \textit{domains}, which describe the symbols that can be used and the possible actions that can be planned over. However, before forming PDDL problems, we first adjust the scene that the agent works with. This serves to speed up planning, as described in section \ref{sec:plan}.

The domain is formed via provided object and action symbols along with manually encoded interactions within the simulator. Primitive actions are qualified with \textit{conditional effects}, which allows one to encode more complex, contextual interactions into the domain; e.g. if a plate is moved underneath a toggled faucet, then the planner should assume its state is changed to \texttt{isClean}. Within the domain one may also encode action costs, which specify priorities for actions.

An example problem, which contains relevant objects, initial conditions, and goal conditions, looks like the following:

{\small
\begin{verbatim}
(define (problem knife_problem)
    (:domain knife_domain)
    (:objects
        Knife_0 - Knife
        DiningTable_0 - DiningTable
    )
    (:init 
        (isVisible Knife_0)
        (isNear Knife_0)
        (isVisible DiningTable_0)
        (isNear DiningTable_0)
    )
    (:goal (isPickedUp Knife_0))
)
\end{verbatim}
}

An example domain, which contains types of objects to plan with, predicates, and action schemas, may look like:

{\small
\begin{verbatim}
(define (domain knife_domain)
	(:requirements 
	    :adl :strips 
	    :typing :conditional-effects
	)
	(:types 
	    Knife - Pickupable
	    DiningTable - Receptacle
	    ... ; more types
	)
	(:predicates
	    (isVisible ?x - Object)
	    (isNear ?x - Object)
	    (isPicked ?x - Pickupable)
	    (parentReceptacles 
	        ?x - Object 
	        ?z - Receptacle
	    )
	    ... ; more predicates
	)
(action Pickup
    :parameters (?x - Pickupable)
    :precondition (and
        (isNear ?x) 
        (forall (?z - Pickupable) 
            (not (isPickedUp ?z))
        )
    )
    :effect (and (isPickedUp ?x)
        ; remove from its parents
        (forall (?z - Receptacle) 
            (not (parentReceptacles ?x ?z)))             
        ; children should all be picked
        ; up if the parent is picked up
        (forall (?z - Object)
            (when
                (parentReceptacles ?z ?x)
                (isPickedUp ?z)
            )
        )
    )
)
... ; more actions
)
\end{verbatim}
}

\paragraph{Subgoal predicates}
Below we describe the subgoal predicates used specifically in our TEACh experiments.
{\small
\begin{itemize}
    \item \texttt{isCooked}: True if a food item has been cooked. Can be the result of being placed in a toaster, a microwave, or on a stove.
    \item \texttt{isClean}: True if an object is clean. Can be the result of being placed under running water.
    \item \texttt{isPickedUp}: True if an object is currently in the agent's hand.
    \item \texttt{isFilledWithLiquid}: True if an item is filled with some liquid, whether water or coffee. Can be the result of being placed under running water or in a coffee machine.
    \item \texttt{isEmptied}: True if an item has nothing inside it.
    \item \texttt{isSliced}: True if an item exists as the result of a slicing operation, such as a bread slice or tomato slice.
    \item \texttt{simbotIsFilledWithCoffee}: True if an object is filled with coffee. Can be the result of an item being placed under a coffee machine.
    \item \texttt{isPlacedTo}: True if an object $x$ currently lies in/on an object $y$.
    \item \texttt{isToggled}: True if an object has been toggled on, such as a light switch or a coffee machine.
\end{itemize}
}

\algnewcommand{\LeftComment}[1]{\(\triangleright\) #1}
\begin{algorithm}[t]
\SetAlgoLined
\KwIn{PDDL domain $D$, Context $C$}
\SetKwFunction{FMain}{HierarchicalPlan}
\SetKwProg{Fn}{Function}{:}{}
\Fn{\FMain{$D$, $C$}}{
    $S ~\leftarrow$ \texttt{InitializeRepr}($C$)\\
    $G ~\leftarrow$ \texttt{SubgoalPredictor}($C$)\\
    \ForEach{$g\in G$}
    {
        \LeftComment{{\color{purple}PDDL Planning}}\\
            $P ~\leftarrow$ \texttt{FormPDDLProblem}($g$, $S$, $D$)\\
            $\Pi ~\leftarrow$ \texttt{PDDLPlanner}($P$, $D$) \\
            \ForEach{$a \in \Pi$}
            {
                $o, e \leftarrow$ \texttt{ExecuteAction}($a$)\\
                \If{$e \neq \texttt{None}$}{
                    \LeftComment{{\color{purple}Error handling}}\\
                    \eIf{$e \in solvable$}{
                        $g' ~\leftarrow$ \texttt{PlanRecovery}($o$)\\
                        \textbf{goto} line 5 with $g:=g'$
                    }{\textbf{return} with failure}
                }
                \LeftComment{{\color{purple}Internal representation update}}\\
                $S ~\leftarrow$ \texttt{UpdateRepr}($o, S$)
            }
    }
    \textbf{return}
}
\textbf{End Function}
\caption{Hierarchical Planning}
\label{alg:online_planning}
\end{algorithm}

\subsubsection{Online Planning Algorithm}
Our full planning algorithm, including the re-planning functionality to recover from failed actions, is listed in Algorithm~\ref{alg:online_planning}.
The subprocesses used in the algorithm are the following:
{\small
\begin{itemize}
    \item \texttt{InitializeRepr}: given a context $C$, initialize the state representation $S$ and return it
    \item \texttt{SubgoalPredictor}: given a context $C$, output a set of predicted subgoals $G$
    \item \texttt{FormPDDLProblem}: given a single goal $g$, a state representation $S$, and a PDDL domain $D$, output a primitive-level plan $\Pi$
    \item \texttt{ExecuteAction}: Given a single primitive action $a$, execute it and return an observation $o$ and error state $e$.
    \item \texttt{UpdateRepr}: given an observation $o$ and state representation $S$, update $S$ with the new observation.
\end{itemize}
}

\subsection{Implementation Details of Baselines}

The architecture of the HET model is shown in Figure \ref{fig:het}. We use two transformer models to implement the controller and the navigator respectively. For the HET-ON model, the navigator is replaced with an oracle navigator which can get access to the ground truth environmental state for reliable navigation. We conduct an extensive hyperparameter search over the model architecture and optimization process. We find that using a 4-layer Transformer with a hidden size of 768 and 12 attention heads obtains the best results on the validation set (average of seen and unseen split). We find the navigator achieves the best performance when increasing the layer number from 4 to 6.

\subsection{\acronym{} GUI Interface}
Figure \ref{fig:visualizer_demo} show a snapshot of the GUI interface we built for visualizing \acronym{}'s intermediate representations.

\subsection{Results on Original Validation Split}\label{apx: original split}
\addafterrebuttal{The original test splits of the TEACh dataset are used for the SimBot Challenge,\footnote{\url{https://eval.ai/web/challenges/challenge-page/1450/overview}} and are thus not publicly available at the time of writing this paper. To facilitate model evaluation, the authors of TEACh re-split the original validation set into new validation and test sets, which are used in this paper as the standard evaluation protocol.
In Table \ref{tb:original_valid}, we additionally provide the results on the EDH subtask of TEACh using the original validation splits. Future work evaluated on the original split can refer to these results when comparing with our model.}

\begin{table}[t]
\centering
\resizebox{.99\linewidth}{!}{
\begin{tabular}{lcccc}
\toprule
\multirow{3}{*}{Model}&\multicolumn{2}{c}{Valid Seen} & \multicolumn{2}{c}{Valid Unseen}   \\ 
\cmidrule(lr){2-3} \cmidrule(lr){4-5} 
& SR & GC & SR & GC \\
\midrule
ET       & 8.55 (0.67) &  9.10 (3.39) &  7.86 (0.91) &  6.20 (3.43)  \\
HET      & 8.72 (1.00) &  9.95 (5.29) &  9.31 (1.74) &  9.50 (5.35)  \\
HET-ON   & 12.01 (3.43) &  16.45 (9.93) &  12.42 (3.43) &  14.95 (9.41)   \\
\addafterrebuttal{JARVIS}   & 15.10 (3.30) & 22.60 (8.70) & 15.80 (2.60) & 16.60 (8.20)  \\
\acronym{}-PaP & 15.62 (7.76) &  20.39 (18.33) &  14.19 (6.12) &  19.32 (17.57)  \\
\acronym{} (Ours)   & \bf 17.76 (9.28) & \bf 24.93 (22.20) & \bf 16.98 (7.24)  & \bf 23.44 (19.95) \\
\bottomrule
\end{tabular}}
\caption{EDH results on the original validation splits.    
}
\label{tb:original_valid} 
\end{table}


\end{document}